\documentclass[11pt,letter]{article}

\newif\iffigures
\figurestrue 

\usepackage{setspace}
\usepackage{amsmath} 
\usepackage{amssymb,nicefrac}
\usepackage{color}
\usepackage{graphicx}
\usepackage[section]{placeins}
\usepackage{pst-node}
\usepackage{dsfont}
\usepackage{mathrsfs}
\usepackage{float}
\usepackage[T1]{fontenc}
\usepackage{aurical}
\usepackage{tikz,pgfplots,verbatim}
\usetikzlibrary{spy,arrows,backgrounds,plotmarks,shapes,snakes}
\usepackage{subcaption}
\usepackage{todonotes}

\usepackage[bookmarks=false]{hyperref}
    \hypersetup{colorlinks,
      linkcolor=blue,
      citecolor=blue,
      urlcolor=blue}
\usepackage{graphicx}
\graphicspath{ {./images/} }
\usepackage[]{algorithm2e}

\usepackage{listings}
\usepackage{color}
\usepackage{soul}

\definecolor{mygray}{rgb}{0.5,0.5,0.5}
\definecolor{mygreen}{rgb}{0,0.6,0}
\definecolor{myorange}{rgb}{0.8,0.4,0}

\newtheorem{proposition}{Proposition}[section]

\definecolor{Yellow}{rgb}{1, 1, 0}
\definecolor{VeryLightGray}{gray}{.90}
\definecolor{LightGray}{gray}{.7}
\definecolor{Gray}{gray}{.50}
\definecolor{DarkGray}{gray}{.3}
\definecolor{VeryDarkGray}{gray}{.10}

\begin{document}

\title{Automated Knowledge Graph Learning in Industrial Processes}

\author{Lolitta Ammann \and Jorge Martinez-Gil \and Michael Mayr \and Georgios C. Chasparis\thanks{Software Competence Center Hagenberg GmbH, Softwarepark 21, A-4232 Hagenberg, Austria; E-mail: {\rm georgios.chasparis@scch.at}.} }

\maketitle

\begin{abstract}
Industrial processes generate vast amounts of time series data, yet extracting meaningful relationships and insights remains challenging. This paper introduces a framework for automated knowledge graph learning from time series data, specifically tailored for industrial applications. Our framework addresses the complexities inherent in industrial datasets, transforming them into knowledge graphs that improve decision-making, process optimization, and knowledge discovery. Additionally, it employs Granger causality to identify key attributes that can inform the design of predictive models. To illustrate the practical utility of our approach, we also present a motivating  use case demonstrating the benefits of our framework in a real-world industrial scenario. Further, we demonstrate how the automated conversion of time series data into knowledge graphs can identify causal influences or dependencies between important process parameters.
\end{abstract}
\section{Introduction}
\label{sec:Introduction}

Industrial processes generate vast amounts of time series data, but turning this data into valuable insights remains a challenging task \cite{Bader}. Our research introduces a new framework that automatically creates knowledge graphs (KGs) from time series data with the help of several established and novel methods. The rationale behind transforming raw data into connected KGs is to help industrial stakeholders optimize processes and facilitate knowledge discovery \cite{NoyGJNPT19}.

Our research introduces a novel approach employing diverse techniques to identify entities and relationships within data, thereby facilitating an interconnected representation of industrial knowledge. The framework developed improves users' understanding of industrial processes and supports the development of advanced modeling techniques, such as causal-inference models for root-cause analysis and predictive models. This framework is among the first to address industrial knowledge management from this perspective, providing a solution for integrating and analyzing complex industrial data.

The proposed framework is based on four primary components: (a) \emph{data pre-processing and quality}, (b) \emph{correlation analysis}, (c) \emph{causality analysis}, and d) \emph{KGs generation}. The initial steps of data pre-processing and quality assurance are standard for industrial sensor time-series data. However, our approach extends beyond these basics by incorporating correlation and causality analysis techniques to identify a broad spectrum of relationships or influences between process parameters. In particular, correlation analysis identifies interrelations or connections in the process parameters' behavior, which may include external factors such as weather conditions influencing input material properties. In addition, causality analysis captures causal influences among process variables over time, which is particularly relevant for dynamic processes. For instance, an operator's control input or setpoint can have a prolonged impact on the final product quality.

The rest of this paper is structured as follows: In the remainder of the Introduction Section~\ref{sec:Introduction}, we provide an overview of related work in KGs and their applications in industrial settings, as well as a discussion on our contributions. Section~\ref{sec:methods} we present the main methodological elements of our framework, detailing the algorithms and techniques used for extracting entities and relationships from time series data. Section~\ref{sec:Demonstrations} presents the results of applying our framework to real-world industrial datasets, showcasing the generated KGs and their utility in specific use cases. Finally, Section~\ref{sec:Conclusions} presents concluding remarks and future work.

\subsection{Related Work}
\label{sec:RelatedWork}

The industrial ecosystems have been transformed by the emergence of collaborative paradigms between human-machine systems, including the Internet of Things (IoT), Internet of Services (IoS), and Cyber-Physical Systems (CPS). This evolution has shifted the focus toward improving human-machine collaboration \cite{hoch2022teaming}. The proliferation of newly connected devices and sensors generates vast amounts of data with significant potential value \cite{Banerjee}. This data can be used to enable on-demand manufacturing, optimize resources and machine maintenance, and improve other logistical operations.

Furthermore, the challenge of bringing KG applications to industrial settings is rapidly evolving, focusing on theoretical frameworks and practical implementations \cite{BuchgeherGGE21}. Recent studies have explored a diverse range of use cases \cite{grangel2018seamless}, demonstrating the versatility and potential of this technology in various industrial scenarios \cite{bachhofner2022automated}.

KGs have been proven effective in representing complex industrial data, facilitating tasks such as process knowledge graph construction \cite{yan2020knowime}, machine parameterization \cite{bachhofner2022knowledge}, material profiling \cite{Zhang}, resource allocation optimization \cite{dombrowski2019knowledge}, and root cause analysis \cite{martinez2022root}. Furthermore, they have been successfully applied in diverse industrial settings, from cognitive manufacturing \cite{liu2022knowledge} to developing industry-specific design tools \cite{meckler2021building}. The literature also shows us the importance of explainable AI (XAI) in industrial decision-making, with KGs playing a crucial role in supporting XAI systems \cite{rozanec}.

In addition, the concept of Granger causality \cite{shojaie2022granger}, which plays a vital role in our framework, has not been deeply explored as an application in industrial settings. Granger causality is a statistical hypothesis test for determining whether one measurement can predict another measurement; for example, a signal X causes a signal Y, then past values of X should contain information that helps predict Y. Therefore, it could be used for detecting outliers \cite{qiu2012granger} or irregular patterns \cite{bahadori2012granger}, for example. In the context of this work, this concept is fundamental because it allows us to gather relevant information about industrial processes and facilitate decision-making around the building of predictive models.

As research and development in this field continue to advance, we can anticipate a future where industrial processes are increasingly monitored and optimized. As more possibilities are explored in terms of managing KGs with machine learning techniques, for example, through the use of KG embeddings \cite{Garofalo}, new applications can be discovered that benefit from having a much more explicit, structured, and interoperable knowledge \cite{rivas2020unveiling}. Therefore, frameworks such as ours are needed to extract and model knowledge with minimal human intervention.

\subsection{Contributions}  \label{sec:Contributions}

We propose a framework for automatically generating KGs from industrial time-series data. Our two main cornerstones of this framework is correlation and causality analysis. In parallel, a modified Granger-causality test with automated time-lag selection is introduced to address the challenge of identifying causal relationships without prior knowledge of process details. The major contributions of this work can be summarized as follows: 

\begin{enumerate}
    \item We introduce a framework that converts raw industrial time series data into structured KGs with minimal manual intervention. This framework employs various techniques to identify and categorize entities and relationships within the data accurately.
    \item  We incorporate the concept of Granger causality, which has not been widely explored in industrial settings. This concept enables the prediction of certain signal behaviors based on other signals, thus moving away from the prevalent black-box models in industrial environments. 
    \item The interconnected KG provides insights into industrial processes, assisting stakeholders in identifying optimization opportunities and making informed decisions. We demonstrate the framework's effectiveness on real-world industrial datasets, showcasing its practical applications and the tangible benefits it offers to industrial stakeholders.
\end{enumerate}

\subsection{Motivating Example} \label{sec:MotivatingExample}

\begin{figure}[th!]
    \centering
    \includegraphics[width=0.8\textwidth]{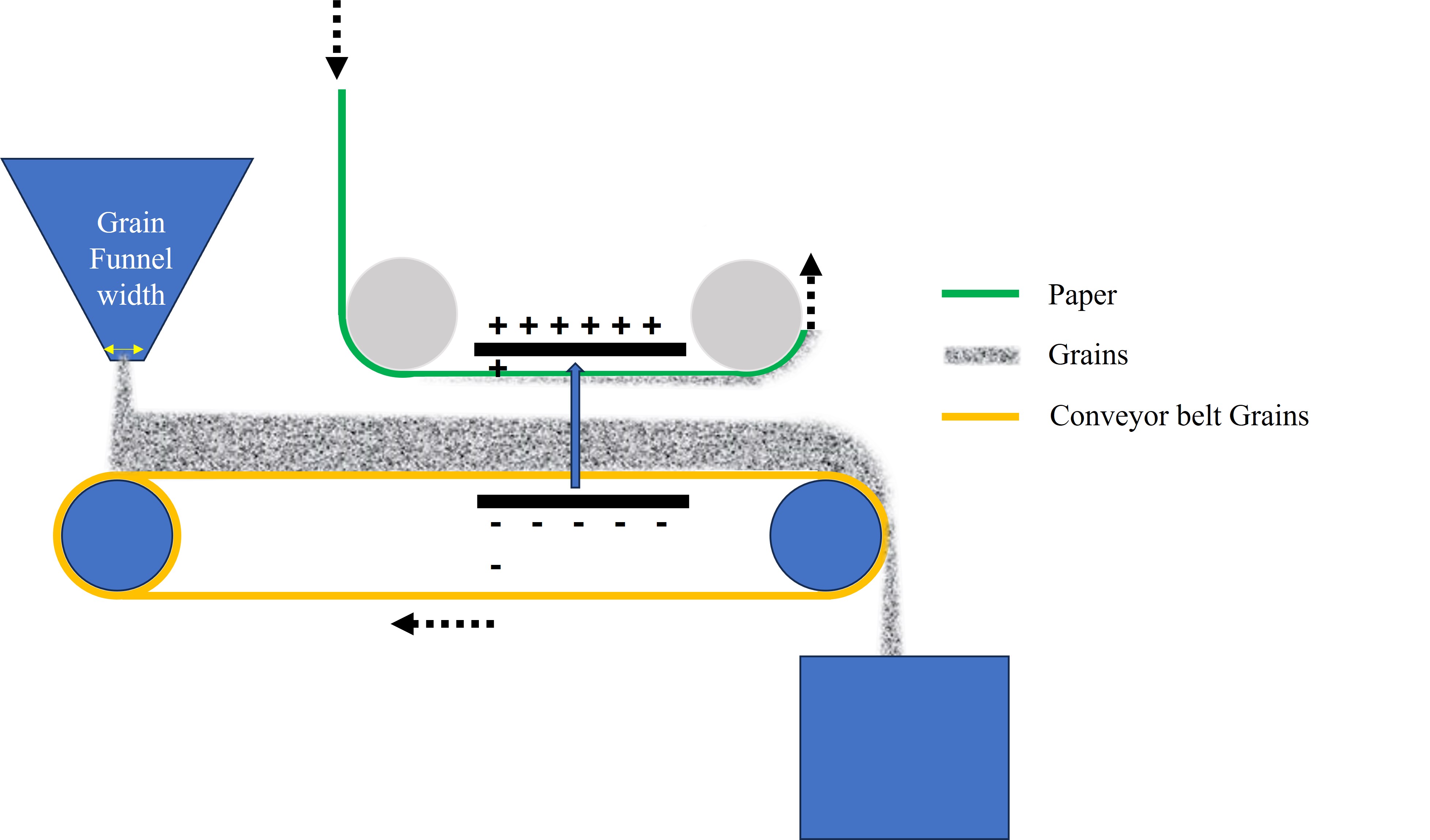}
    \caption{Sketch of a real-world industrial process of electrostatic particle transfer.}
    \label{fig:UseCase}
\end{figure}

To highlight the need for such a framework of automatically generating KGs that captures relationships between process parameters, we consider a real-world industrial use-case, commonly referred to as \emph{electrostatic particle transfer}. It finds applications to different types of industries (e.g., dust removal \cite{panat_electrostatic_2022}, electrostatic dry powder coating in pharmaceutical industry \cite{vanamu_overview_2022}, paper coatings \cite{vaha-nissi_coating_2016}, abrasive powders \cite{tkach_electrostatic_2021}). A typical setup of such process is depicted in Figure~\ref{fig:UseCase} that describes the functionality of the electrostatic field in the case of the paper coating industry. In this case, the grains or powder is supplied to a conveyor belt through a funnel with an adjustable opening that regulates the flow of material. The conveyor belt moves the material through a static electric field formed by two parallel plates. The particles are first electrically charged (with the negative charge) and then attracted to the positively charged plate by electrical forces. In the case of paper industry, the paper has already been coated with a special glue that helps the material to stick to the paper against the gravitational forces. The goal of the process is to create a desirable density of material (grains/powder) on the paper through the proper regulation of several configuration parameters, such as the flow of material, the current/voltage/frequency of the electric field, the distance of the plates at the beginning and at the end and the speed of the conveyor belt.

Unfortunately, there are several challenges that makes the correct calculation of the configuration parameters to achieve a desirable density. For example, the humidity of the particles may not necessarily be a-priori known to the operators, or the size of the particles may influence differently the movement of the particles, or aerodynamic phenomena may take place that cannot be predicted in detail. Such exogenous influences prohibit the use of exact physics-based mathematical formulations that can accurately calculate the resulting density of material on the paper. This gives rise to the need of data analysis techniques to understand the influence of the different configuration parameters, and also accurately predict the resulting material density on the paper. Furthermore, in cases where abnormal behavior is observed, it is useful for the operator to be informed regarding the parameter or the combination of parameters that have most likely led to the observed phenomena. These type of questions have motivated the generation of automated KG discovery.

\section{Methods}
\label{sec:methods}

\subsection{Proposed Framework} \label{sec:Framework}

\begin{figure}[th!]
    \centering
   \includegraphics[scale=0.3]{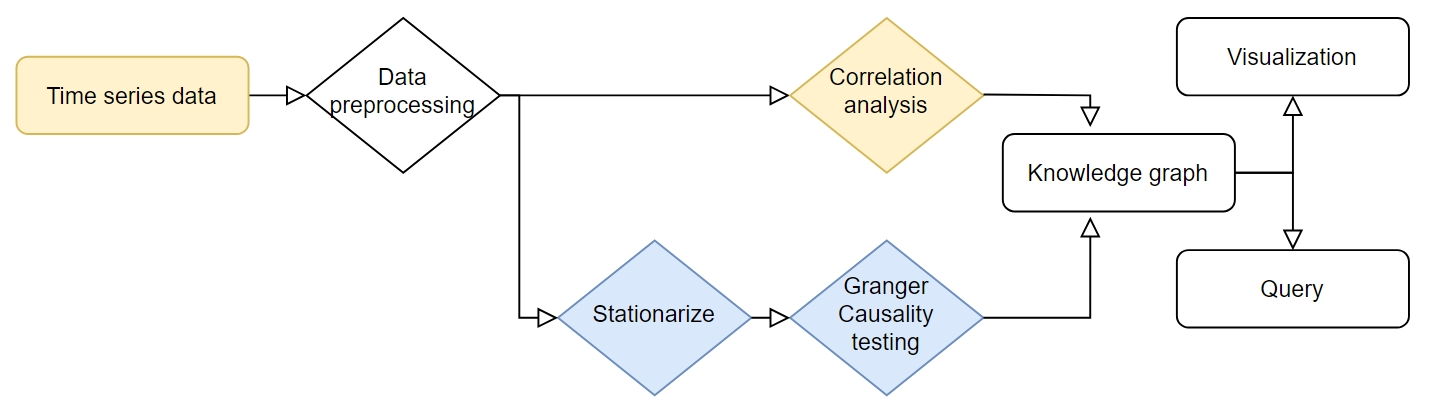}
    \caption{A general overview of the proposed framework for automated KG discovery.}
    \label{fig:Framework}
\end{figure}

Our research focuses on time series data generated by sensors in industrial processes. This data can help identify patterns, detect anomalies, predict maintenance needs, and improve overall process efficiency. We have developed a framework to integrate these capabilities that facilitate data collection, pre-processing, analysis, visualization, and decision-making support. It is designed to manage static and real-time data streams, offer analytics, and provide intuitive dashboards for end-users. The framework includes scalable data storage solutions, supports the development of advanced anomaly detection algorithms and predictive maintenance models, and features user-friendly interfaces for easy interaction. A rough sketch of the functionality of the proposed framework is depicted in Figure~\ref{fig:Framework}.

The framework is implemented as a web application, enabling users to upload sensor data, select specific sensor measurements for analysis, and choose methods for numeric conversion and missing value imputation. It offers various options for selecting these parameters and includes several methods to compare selected measurements, such as Pearson correlation, Spearman rank correlation, and Euclidean similarity. Additionally, the framework can easily accommodate other comparison methods. Furthermore, it provides a Granger causality testing methodology for detecting causal influences between process parameters (including the direction and delay of causal effects).

The developed application supports KG creation in RDF format and querying using the SPARQL language to explore correlations and causations within the data. This improves the interpretability of industrial processes, moving beyond the traditional black box approach where raw data offer limited insights.

The visualization functionality facilitates in-depth analysis, while integrated models provide predictive insights based on the sensor measurements. The framework's flexibility allows for customization and extension, making it suitable for various research and industrial applications.

In the following sections, we describe in more detail the three main methodological elements of our framework that leads to the automated generation of KGs.

\subsection{Pre-processing}

Pre-processing is a crucial step before conducting our correlation and causality analysis. The idea is to ensure the quality and reliability of the results by addressing potential issues in the raw sensor data. Sensor data often has missing values due to sensor malfunctions, transmission errors, or other reasons. Our framework offers imputation methods (filling in missing values) like mean imputation, median imputation, or other more advanced techniques. The choice depends on the nature of the data and the potential impact of missing values on the analysis. The management of categorical sensor data is also possible. The reason is that sometimes, there is a need to encode these data into numerical values before performing correlation analysis. 

\subsection{Correlation analysis}   \label{sec:CorrelationAnalysis}

\begin{figure}[t!]
    \centering
   \includegraphics[scale=0.6]{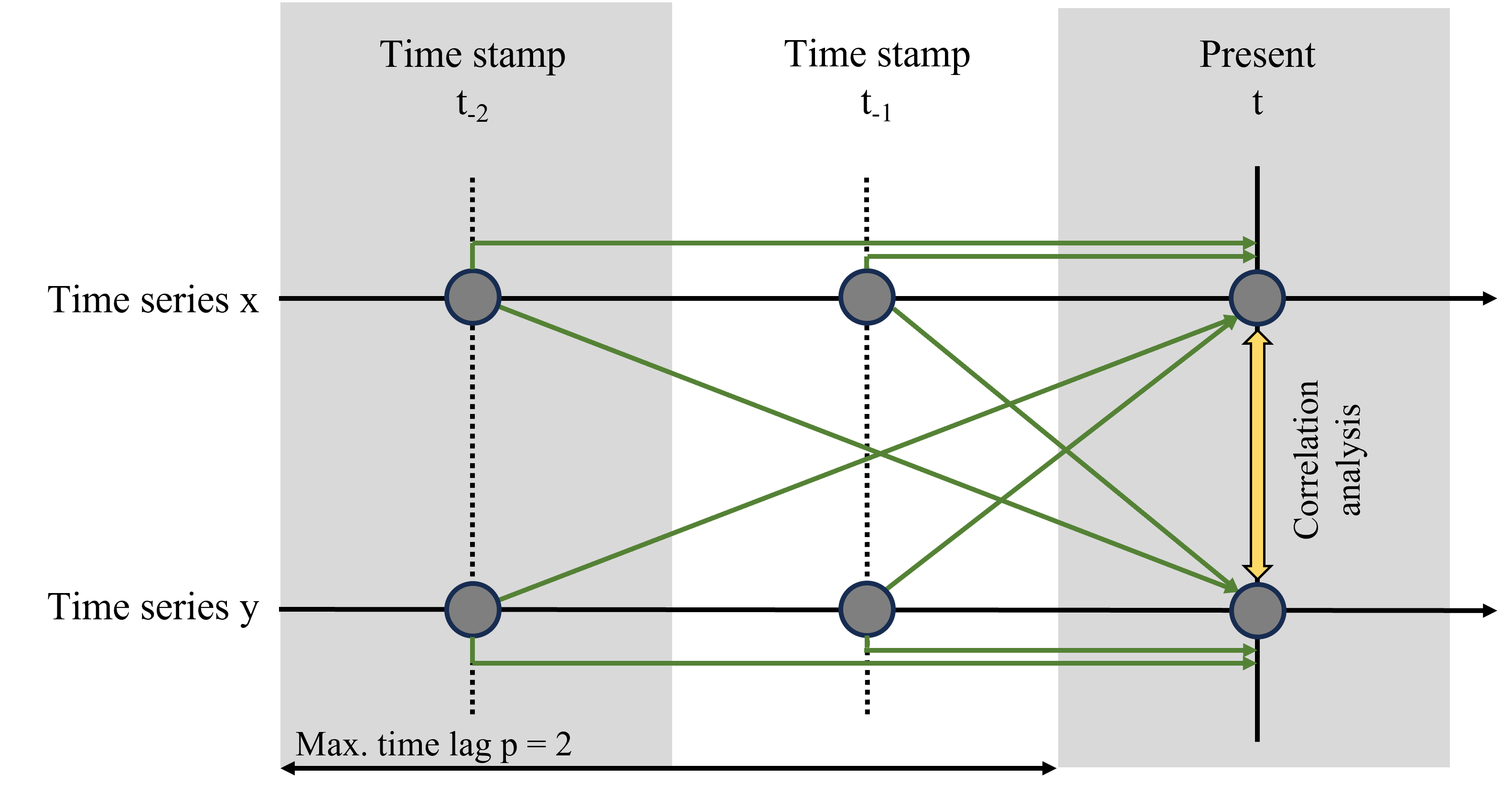}
    \caption{Combination of correlation analysis and Granger causality for search of contemporary and lagged relations between parameters. Green arrows depict inspected connections using Granger causality on VAR($p$) model, where $p$ corresponds to the time-lag over which causality is tested. The yellow arrow depicts contemporary correlation analysis.}
    \label{fig:CorrelationCausality}
\end{figure}



Correlation analysis examines the strength and direction of relationships between two or more random variables within the same time interval, as depicted in Figure~\ref{fig:CorrelationCausality}. In this context, these random variables may represent  different sensor measurements (e.g., temperature, humidity, pressure). We employ several types of correlation metrics, including the following: 
\begin{itemize}
    \item \emph{Pearson Correlation.} Measures the linear relationship between two continuous variables. High positive correlation means they tend to increase or decrease together. High negative correlation means one increases while the other decreases.
    \item \emph{Spearman Rank Correlation.} Assesses the monotonic relationship between two variables (not necessarily linear). This is useful when dealing with data that does not strictly follow a normal distribution.
    \item \emph{Euclidean Similarity.} Measures the geometric distance between data points. Closer points are more similar. This can be helpful for clustering or finding patterns.
\end{itemize}

In the first version of the framework, we have decided to add these measures because in our experience they tend to be some of the most useful, but adding new comparison methods is straightforward. It is worth noting that the Pearson correlation analysis requires normality of the investigated time-series processes. In most cases, a graphical test is sufficient to check such condition.

\subsection{Granger Causality Analysis}     \label{sec:GrangerCausalityAnalysis}

Granger causality, originally developed for economic applications \cite{granger_investigating_1969}, is increasingly used in various fields to identify causal-effect relationships between time series data. A special perk is the ability to hypothesize about the influence of one time series on another including an estimate of the delay in the effect (\emph{time lag}) and consequently, the direction of the relation. This is investigated by testing for improvement of a model by including previous time stamps of additional time series. For two stochastic processes, $x_t$ and $y_t$, if the forecast predictor of $x_t$ is improved by including all current and prior values of $y_t$, then we say that $y_t$ is \emph{Granger-causal} for $x_t$. 

The concept of Granger-causality is not restricted by the type of forecast model, however it is straightforward to check it in the context of vector auto-regressive model (VAR). Here, we estimate the full VAR model and test for a significant increase in variance upon exclusion of specific time series or specific time lags using F-tests, i.e. we test for Granger non-causality. Below we describe the approach in more detail starting with how the VAR model is used and estimated in this particular case.

\subsubsection*{VAR($p$) model and Granger-Noncausality}

A VAR($p$) model predicts the next values of a time series using the information available within the last $p$ time intervals. Let us consider two one-dimensional real-valued time series $x_t$ and $y_t$ that are modeled (or predicted) using a VAR($p$) model as follows:
\begin{subequations}    \label{eq:ar_model}
\begin{align}    
    x_t & =  \nu_{x} + \sum_{i=1}^{p} \alpha_i x_{t-i} + \sum_{i=1}^{p} \beta_i y_{t-i} + u_{x,t}\,, \quad t=0,\pm1,\pm2,\ldots \\
    y_t & =  \nu_{y} + \sum_{i=1}^{p} \gamma_i x_{t-i}+ \sum_{i=1}^{p} \delta_i y_{t-i} + u_{y,t}\,, \quad t=0,\pm1,\pm2,\ldots
\end{align}
\end{subequations}
where $\nu_x$ and $\nu_y$ are the intercept terms, and $\alpha_i$, $\beta_i$, $\gamma_i$ and $\delta_i$ are real constants. Also, $u_{x,t}$ and $u_{y,t}$ are white noises with nonsingular covariance matrix $\Sigma_{u,x}$ and $\Sigma_{u,y}$, respectively. 

Consider a new random $z_t=(x_t,y_t)$ and let us denote by $z_t(h|\Omega_t)$ the $h$-step predictor of $z_t$ given all available information up to time $t$, $\Omega_t$. The following proposition can be used for testing Granger noncausality, according to \cite[Corollary~2.2.1]{lutkepohl_new_2005}.

\begin{proposition}[Granger Noncausality]   \label{Pr:Noncausality}
    If $z_t$ is a stable VAR($p$) process\footnote{The stability condition of a VAR($p$) process can be verified by the eigenvalues of its coefficients matrix, cf.~\cite[Section~2.1]{lutkepohl_new_2005}. Stability also implies stationarity.} with a nonsingular white noise covariance matrix $\sigma_u$, then
    \begin{equation}
        x_t(h|\{z_s|s\leq{t}\}) = x_t(h|\{x_s|s\leq{t}\})\,, \quad h=1,2,... \Leftrightarrow \quad  \beta_i = 0\,, \quad i = 1,...,p
    \end{equation}
\end{proposition}

This proposition is the basis of our investigation, since it implies that noncausalities can be determined by just looking at the VAR($p$) representation of the time-series of interest.

\subsubsection*{Estimating a VAR($p$) model}

Investigating noncausality between two time-series of interest implies the estimation of a VAR($p$) process. In this paper, we are using \emph{multivariate linear-squares estimation} (LS) for formulating these estimates. It is considered multivariate since estimates need to be formulated for pairs or sets of random variables, as Proposition~\ref{Pr:Noncausality} requires. For example, we can test when adding a new variable to the existing set of variables can improve the investigated estimator. Computing an LS estimator is considered rather standard and the interested reader may refer to \cite[Section~3]{lutkepohl_new_2005} for a detailed description of its solution. 

\subsubsection*{Testing for Granger-noncausality}



Let us consider that we have collected measurements of $z_t=(x_t,y_t)$ over a time horizon of $T$ time steps. The proposed approach takes as inputs: 1) a row vector $Z$ of length $K(T-p+1)$, containing the set of $K$ time-series variables under investigation over a time lag $p$ (e.g., in the example of Equation~(\ref{eq:ar_model}) that would be the pair of  $x_t$ and $y_t$) with the LS estimate being denoted by $\hat{Z}$; 2) a row block matrix $A$ to be filled with the regression coefficients, and 3) the data ($x_{t-i}$ and $y_{t-i}$) rearranged in matrix form in such a way that allows to compute the dot product of $A$ and the data and retain a row vector of estimates $\hat{Z}$. The model is estimated using OLS with an SCS solver from the python cvxp library.


As evident from Proposition~\ref{Pr:Noncausality}, this framework allows for testing for Granger noncausality in a very flexible way. For example, in the case the full model consists of two random variables, as in Equation~(\ref{eq:ar_model}), we compare the full model with the model where the parameters of interest are excluded by setting the respective regression coefficients to zero (according to Proposition~\ref{Pr:Noncausality}). In particular, if we set $\beta_i\equiv{0}$, we compare with a model where the evolution of $x_t$ does not depend on $y_t$ for a certain history length $p$. 
If the model with the excluded parameters is not worse, the excluded parameters did not lead to an improvement of the model and a relation between the parameters is not expected. The exclusion is induced in the computation of LS optimization of the regression parameters with additional constraints.


An F-test allows comparison of the full model with the constrained model, which is based on computing the following index value:
\begin{equation}
F = \left(\frac{SS_{c} - SS_{f}}{p}\right)\Big/\left(\frac{SS_{f}}{T-p+1}\right)
\end{equation}
where: ${SS}_c$ is the sum of squares of the constrained model (where we set the corresponding regression coefficients to zero), ${SS}_f$ is the sum of squares of the full model, and $p$ is the considered time lag. 

\subsubsection*{Assumptions}


Testing for Granger noncausality requires assumptions such as stationarity of the processes. Stationarity is a property that ensures constant means, variances, and autocovariances of the process through time \cite{lutkepohl_new_2005}. There is a range of tests available to check for different types of non-stationarity. A commonly used approach is the Dickey-Fuller test, which checks for unit roots in the process. If non-stationarity identified in the process the data is differenced, i.e. rather than describing the time series by the actual values, it is described by the change from timestamp $t$ to the timestamp $t+1$. This process can be repeated several times, with the order of integration describing how many times the data had to be differenced to retain stationarity. Test statistics such as the $F$ test used to compute Granger causality can suffer from loosing the asymptotics properties of the $\chi^2$ distributions in the case of differenced data. While this is not an issue in the current version of the tool, where never all lags are constrained in the same model, this can become an issue, if effects of time series are tested irrespective of time lag, i.e. all lags are constrained. Following the procedure above for a var(p + 1) model can overcome this issue, assuming the additional lag does not improve the model and thus will not include constraints.  

\subsection{Query capabilities}
Representing multi-dimensional time-series data and their correlations/causations in a Knowledge Graph enables efficient and intuitive visualizations and sub-graph filtering using graph-oriented query languages (e.g. SPARQL\footnote{\url{https://www.w3.org/TR/sparql11-query/}} for RDF graph data formats, Cypher for Neo4J\footnote{\url{https://neo4j.com/}}, or AQL for ArangoDB\footnote{\url{https://arangodb.com/}}). This approach facilitates in-depth analysis, allowing users to perform complex queries over KGs, simplifying the retrieval of, e.g. lagged co-variates that causally impact specific targets given certain thresholds, enabling threshold-based filtering to retrieve and visualize sub-graph structures. In more complex settings, users may query the KG to find all sensors that showed a temperature increase above a certain threshold before a specific event or correlate these changes with other environmental factors. Additionally, utilizing semantic reasoning may infer new knowledge from existing data. For instance, if certain conditions are known to precede equipment failures, the system can infer potential risks before they occur.


\section{Demonstrations}    \label{sec:Demonstrations}

\begin{figure}[th!]
    \centering
    \includegraphics[scale=0.8]{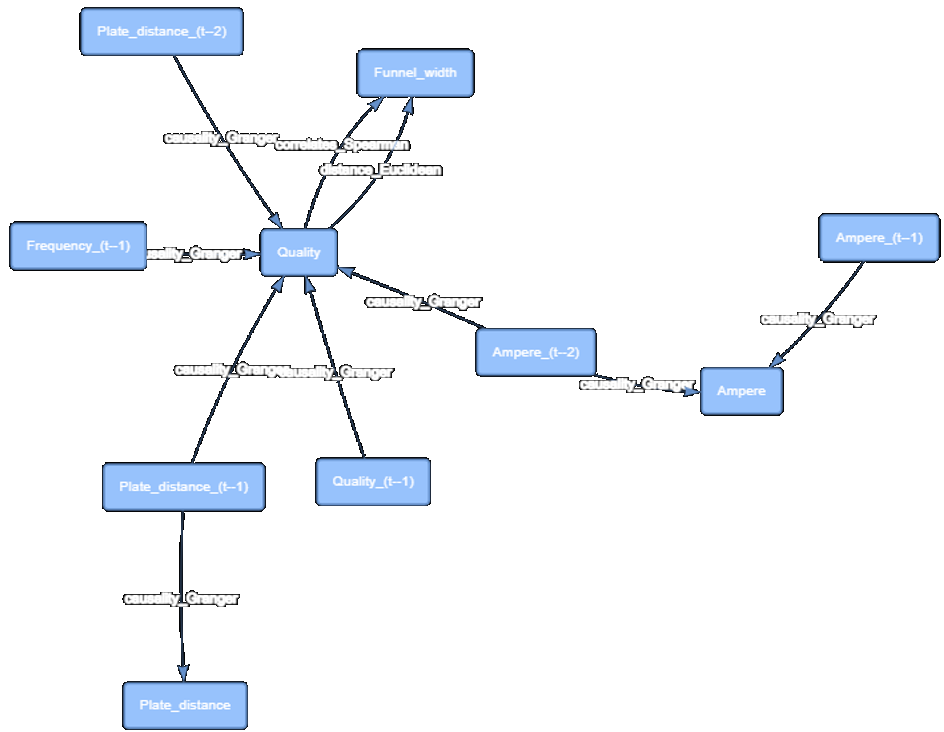}
    \caption{Example of automated KG generation based on correlation and causality analysis in the electrostatic particle transfer industrial process of Section~\ref{sec:MotivatingExample}. }
    \label{fig:CorrelationCausalityGraph_Electrostatic}
\end{figure}

We employed the proposed correlation and Granger-causality discovery described in Sections~\ref{sec:CorrelationAnalysis}--\ref{sec:GrangerCausalityAnalysis} to the real-world industrial process of Section~\ref{sec:MotivatingExample}. This leads to the generation of a KG that represents both instantaneous correlations between process parameters as well as time-lagged causal influences. The resulting KG is depicted in Figure~\ref{fig:CorrelationCausalityGraph_Electrostatic}. In the case of the electrostatic particle transfer, the target or predicted variable is the quality of the coating process (which represents the density of the powder/grains on the paper).

The KG of Figure~\ref{fig:CorrelationCausalityGraph_Electrostatic} shows reasonable outcomes according to expert knowledge, since it indeed demonstrates the expected causal and correlation dependencies. For example, as expected, the plate distance, the frequency and strength of the electric field as well as the previous quality of specific time lags, causally influences the upcoming quality. Furthermore, the funnel width is instantaneously correlated with the quality (which is implied to be bidirectional). 

\begin{figure}[th!]
    \centering
    \includegraphics[width=0.85\linewidth]{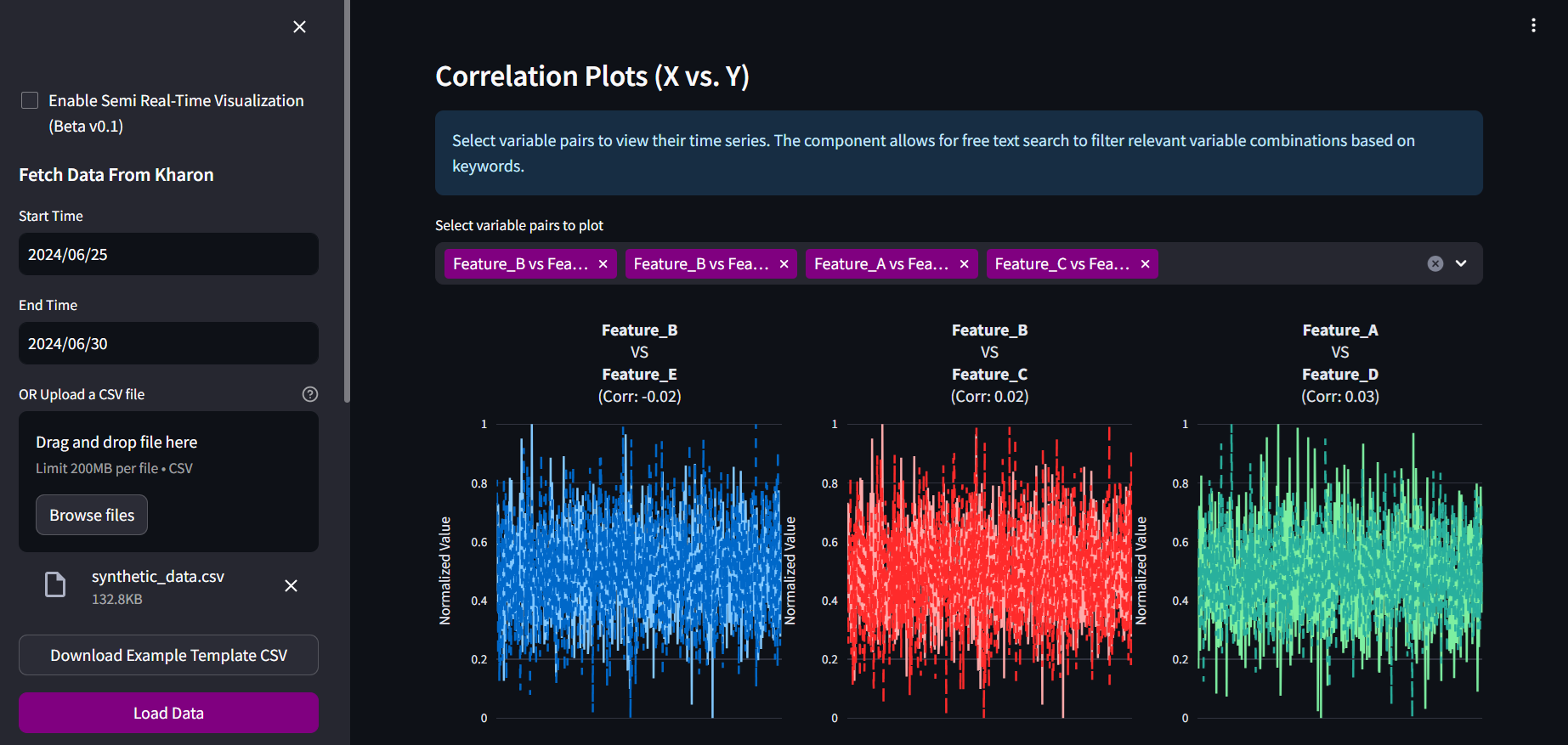}
    \caption{Correlation analysis example in a synthetic data set.}
    \label{fig:CorrelationAnalysis_Example}
\end{figure}
\begin{figure}[th!]
    \centering
    \includegraphics[width=0.85\linewidth]{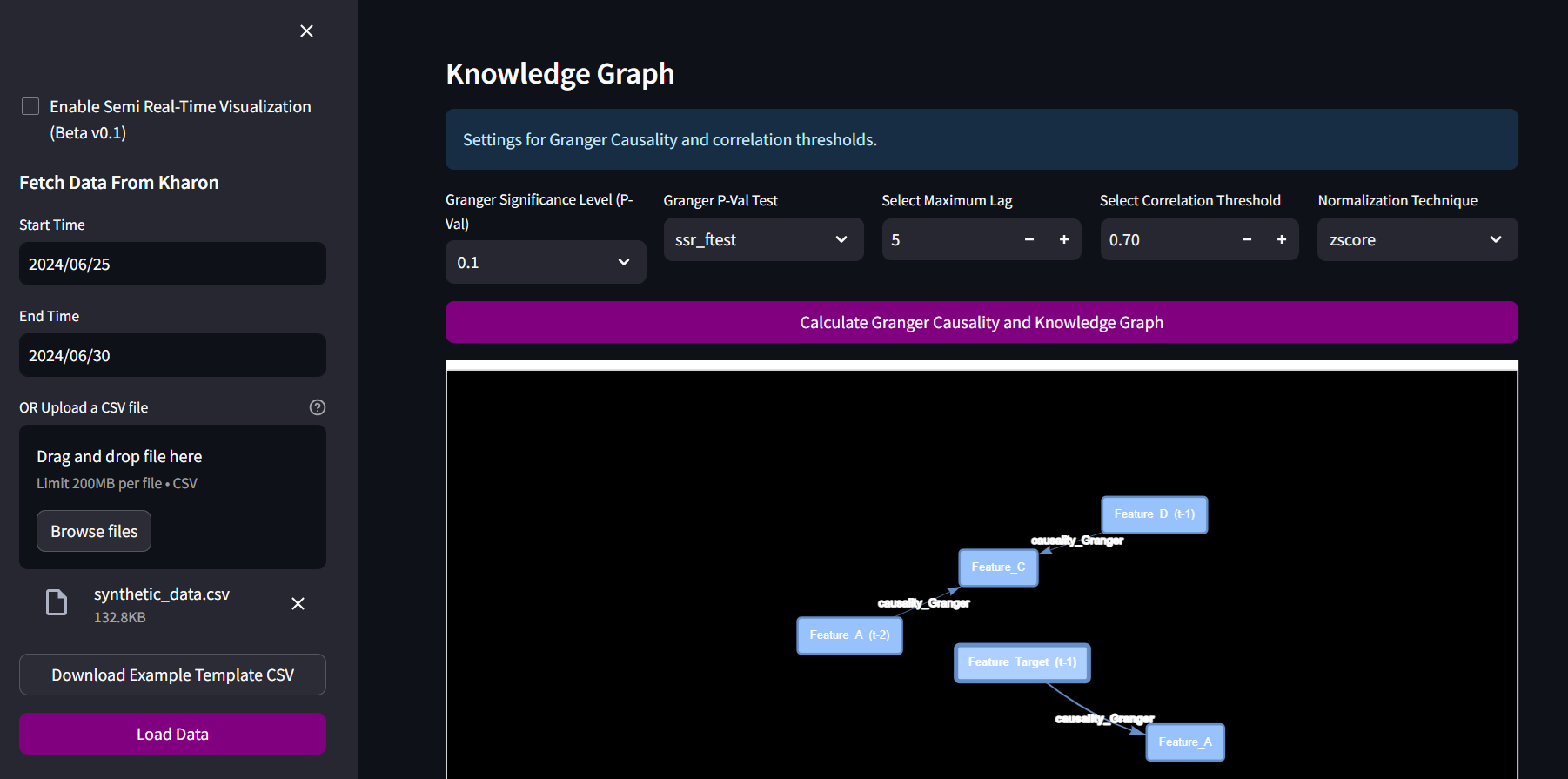}
    \caption{Causality analysis example in a synthetic data set and KG generation.}
    \label{fig:CausalityAnalysis_Example}
\end{figure}

We also designed and implemented a KG application tool that incorporates several functionalities and methods for the three main pillars of our framework, namely preprocessing, correlation and causality, as described in Section~\ref{sec:methods}. In particular, our application includes the possibility of integrated different types of data sources (e.g., connection to a datalake, or uploading CSV files), preprocessing the data, plotting correlations between selected process variables, and finally presenting the resulting KG which also includes the causal dependencies. Below, we show an example of the correlation analysis in Figure~\ref{fig:CorrelationAnalysis_Example}, and the corresponding KG in Figure~\ref{fig:CausalityAnalysis_Example}. 

\section{Conclusions and Future Work}
\label{sec:Conclusions}

This work shows the significant potential of automated KG learning tools in industrial settings. The rationale behind converting time-series data generated through sensor measurements into KGs is to effectively address the challenges of extracting meaningful insights from large datasets. Moreover, applying Granger causality within this framework provides a method for uncovering causal relationships between sensor measurements. Granger causality is helpful in time-series analysis as it identifies potential causal links, rather than mere correlations, between different sensor measurements over time. This capability is crucial for predictive maintenance, anomaly detection, and optimizing industrial processes, as it identifies root causes and predicts future events based on historical data.

The practical use case presented illustrates how this framework can improve decision-making, optimize processes, and facilitate knowledge discovery, eventually leading to improved operational efficiency. Moreover, integrating advanced analytics could facilitate industries' full potential use of their data, facilitating a competitive advantage.

Future work will concentrate on sharpening the framework's capabilities for various industries and further exploring its benefits through additional case studies. We want to improve scalability, integrate the tool with existing KG databases, refine the user interface, and quantify its economic impact. These efforts aim to strengthen the tool's capabilities and broaden its applicability.

\section*{Acknowledgments}
The authors thank Richard Kueng for suggesting constrained least squares minimization using cvxpy. All authors acknowledge financial support by a) the European Union’s Horizon Europe research and innovation program under the ``TRINEFLEX" project (Grant No. 101058174), b) the FFG (www.ffg.at) under the “COGNERGY“ project (Grant No. FO999899014), and c) the Federal Ministry for Climate Action, Environment, Energy, Mobility, Innovation, and Technology (BMK), the Federal Ministry for Digital and Economic Affairs (BMDW), and the State of Upper Austria in the frame of SCCH, a center in the COMET - Competence Centers for Excellent Technologies Programme.

\nocite{*}

\bibliographystyle{abbrv}
\bibliography{mybib}
\end{document}